\documentclass[letterpaper, 10 pt, conference]{ieeeconf}  

\IEEEoverridecommandlockouts                              

\usepackage[T1]{fontenc}
\usepackage[nomath]{lmodern}
\usepackage[table]{xcolor}
\usepackage{url}
\usepackage{amsmath,amssymb,latexsym}
\usepackage{subcaption}
\usepackage{siunitx}
\usepackage{import}
\usepackage[stable]{footmisc}
\usepackage{tikz}

\usepackage[left=17mm,right=17mm,top=21.2mm,bottom=15.2mm]{geometry}

\usetikzlibrary{shapes,arrows,calc,math,arrows.meta}
\title{\LARGE \bf
UPSLAM: Union of Panoramas SLAM
}

\author{Anthony Cowley, Ian D. Miller and Camillo Jose Taylor
\thanks{The authors are with the GRASP Laboratory, School of Engineering and Applied Sciences,
        University of Pennsylvania, Philadelphia PA 19104.}%
\thanks{Corresponding author: 
        {\tt\small acowley@seas.upenn.edu}}%
}

\begin{document}

\maketitle
\thispagestyle{empty}
\pagestyle{empty}

\begin{abstract}
We present an empirical investigation of a new mapping system based on a graph of panoramic depth images. Panoramic images efficiently capture range measurements taken by a spinning lidar sensor, recording fine detail on the order of a few centimeters within maps of expansive scope on the order of tens of millions of cubic meters. The flexibility of the system is demonstrated by running the same mapping software against data collected by hand-carrying a sensor around a laboratory space at walking pace, moving it outdoors through a campus environment at running pace, driving the sensor on a small wheeled vehicle on- and off-road, flying the sensor through a forest, carrying it on the back of a legged robot navigating an underground coal mine, and mounting it on the roof of a car driven on public roads. The full 3D maps are built online with a median update time of less than ten milliseconds on an embedded NVIDIA Jetson AGX Xavier system.
\end{abstract}

\section{INTRODUCTION}
Simultaneous localization and mapping (SLAM) is a core capability underpinning general field robotics. It is a keystone for software needing to relate instantaneous readings from a robot's sensors to historical data -- anything from what the robot perceived a fraction of a second in the past, to a pre-loaded map of a location -- while allowing the robot to extend its corpus of knowledge about how its sensors perceive its surroundings. Relating current sensor data to a map permits localization, while computing transformations that explain how sensor data changes over time enables the extension of that map with new data. The degree to which existing techniques meet the challenge of SLAM depends upon factors such as \cite{cadena2016}: 

\begin{itemize}
\item \textbf{Motion} Will the sensor be moved quickly or slowly? Over smooth or bumpy ground, or through the air?
\item \textbf{Geometry/Appearance} Does the target environment offer enough essential features (e.g. planes, corners, textures)?
\item \textbf{Performance} Accuracy, computational efficiency, and how the approach scales with environment size.
\end{itemize}

Existing mapping systems tend to work well when operating under conditions similar to those in which their original authors demonstrated their performance. But in exploration tasks, our robots reliably discover vantage points with wildly varying feature availability while experiencing rapid, unexpected motions as they traverse unfamiliar terrain. Bumping and bouncing, slipping and sliding, or executing rapid pivots all have the potential to momentarily interfere with motion estimation by reducing overlap between consecutive sensor readings, or by breaking assumed motion priors used to improve optimization convergence (e.g. planar motion). The environment itself plays a role as transitions between enclosed and open spaces, or rounding a corner with poor visibility can lead to rapid significant changes in the types of visual and geometric features captured by a sensor. The challenges of these transient, yet frequently encountered, conditions may be diminished by ensuring that every scrap of information a sensor produces is leveraged by the SLAM system.

Consequently, we investigate a new SLAM method built to make use of as much range data as possible. The robustness of the approach is demonstrated along all three axes identified above by applying it to data collected by a sensor carried by a walking human, a running human, a small wheeled robot platform, a legged robot platform, an aerial robot platform, and an automobile, in environments from indoor clutter to underground tunnels to wild forests to public roads. Its performance is quantified, demonstrating trajectory error of 0.05\% the trajectory length, its ability to run within a \SI{15}{W} power budget, and its scaling from representing the geometry of window mullions to coping with 21.5 million cubic meters of space attached to a \SI{4.4}{km} trajectory.

In this work, the environment is represented as a series of panoramic depth maps. This choice allows us to represent our local maps using a 2D array based representation. Our implementation is able to effectively exploit the regular memory access patterns and parallelism that this representation affords. As a result it runs very quickly even on embedded GPU platforms. An important consequence is that we are able to make use of all available range measurements and avoid the need for early decisions about features. We argue that using all of the data provides additional robustness which allows the method to handle a wide variety of environments.

\section{Related Work}
The SLAM problem has among the deepest wealth of research of any area of robotics. We draw particular inspiration from the idea of building an atlas \cite{Bosse03anatlas} of local maps linked by rigid body transformations. As lidar sensors increase in resolution, keyframe techniques \cite{Meilland13b, meilland:jfr2015} more commonly applied to camera-like sensors gain appeal. With such techniques, an instantaneous data capture serves as the initial value of a local map. In particular, we advocate the use panoramic image keyframes \cite{Taylor:DepthPanoramas:2015} for their suitability to capture data gathered while rotating.

\newgeometry{top=20.1mm,bottom=15.2mm,right=17mm,left=17mm}

Modern geometric SLAM systems, such as LOAM \cite{zhang14_loam}, have found success from winnowing the input data streams of high-resolution lidar through feature selection. In subsequent works, the authors improve robustness by adding complementary sensing modalities as fallbacks for when a primary sensor, typically a lidar, is unable to maintain continuous tracking \cite{zhang18_lvio}. LeGO-LOAM \cite{legoloam2018} is a notable specialization of LOAM to ground vehicle scenarios. It incorporates changes to point cloud filtering and feature selection that enable it to run at the sensor's update rate by tracking fewer features than the original LOAM formulation.

A parallel trend in mapping is driven by the great success found in the use of features detected by visible light and infrared sensitive cameras, with ORB-SLAM \cite{MurArtal:ORB-SLAM:2015} a notable contributor to the popularization of monocular SLAM. It is able to extract and track features quickly enough for motion estimation, while the features it finds are also suitable for loop closing. The advantages of cameras are that they are inexpensive, mechanically robust, and historically endowed with more generous, or at least symmetric, fields of view than laser range finders. However, a limitation of many texture-based feature mapping systems is that they produce a sparse 3D reconstruction of a workspace as defined by the tracked features. This limitation is somewhat alleviated by direct methods such as Direct Sparse Odometry \cite{Engel:DSO:2018} that optimize differences in image intensity values rather than extracted features. Such techniques can produce denser models than feature-based approaches since direct methods leverage more of the available image data. Work combining the dense 3D scanning of RGB-D sensors with more discriminative ORB features has been shown to work for moderately sized scenes\cite{Liu:KeyframeSLAM:2017}.

\section{UPSLAM}

\begin{figure}
  \centering
  \subcaptionbox*{}{\includegraphics[width=0.9\columnwidth]{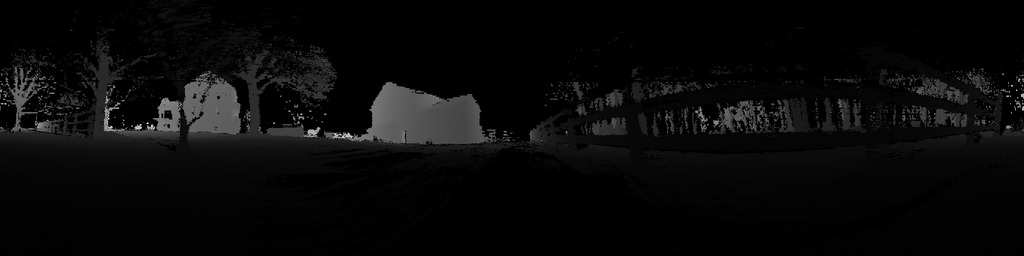}}
  \subcaptionbox*{}{\includegraphics[width=0.9\columnwidth]{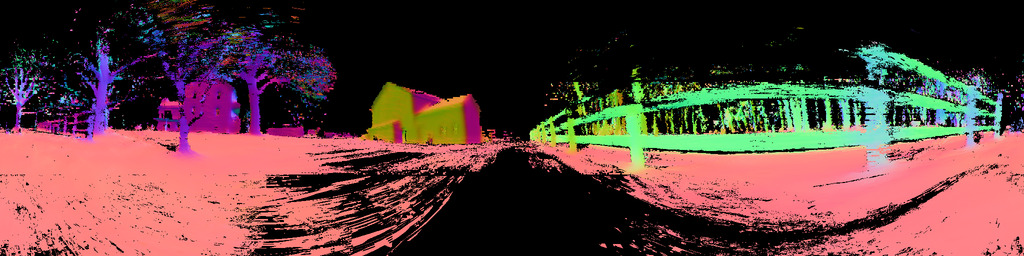}}
  \caption[Farm keyframe]{Example keyframe panorama from a farm lane bordered by a fence. The upper image shows depth information, with lighter shades corresponding to greater depth, while the lower figure maps surface normal vector components to color.}
  \label{fig:farm-keyframe}
\end{figure}

Union of Panoramas SLAM (UPSLAM) represents a map as a graph whose nodes are panoramic depth images augmented with surface normal estimates, Figure~\ref{fig:farm-keyframe}, and whose edges are 3D transformations between those depth image keyframes. Keyframes may have different resolutions and extents than raw sensor data; we typically operate with a wider vertical field of view for keyframes than the sensor captures. The system keeps track of a current keyframe which is updated with each new lidar sweep. The experiments described here were conducted with an Ouster OS1-64 lidar, in which an array of 64 lasers covering a 33° vertical field of view are swept about an axis at \SI{10}{Hz}, and an IMU provides accelerometer and gyroscope measurements at \SI{100}{Hz}.

\subsection{Data Intake}
During rapid motions, the \SI{100}{ms} period of a lidar sweep may include significant sensor motions. Subsequent data processing will search for a rigid body transformation between 3D points derived from a single sweep and a model point set. The tension between applying a single transformation to points collected at different sensor poses is mitigated by using high-frequency inertial measurements to rotate all range measurements to the rotational frame of a single point in time. The rotations based on IMU and lidar timing data are computed on the CPU, while virtually everything that follows is computed on the GPU.

The points resulting from this IMU motion correction are projected into a panoramic image. The image structure is helpful to compute a rough surface normal estimate as the cross product between the vectors defined by a central point and two of its image coordinate neighbors. These surface normal estimates, which are again represented as an image, are very noisy, but may be efficiently smoothed using an edge sensitive smoothing algorithm such as the à trous wavelet filter \cite{Dammertz:ATrous:2010}.

\subsection{Motion Estimation}
\begin{figure}
  \centering
\begin{tikzpicture}[x=1pt,y=1pt]
\definecolor{fillColor}{RGB}{255,255,255}
\path[use as bounding box,fill=fillColor,fill opacity=0.00] (0,0) rectangle (216.81,101.18);
\begin{scope}
\path[clip] (  0.00,  0.00) rectangle (216.81,101.18);
\definecolor{drawColor}{RGB}{255,255,255}
\definecolor{fillColor}{RGB}{255,255,255}

\path[draw=drawColor,line width= 0.6pt,line join=round,line cap=round,fill=fillColor] (  0.00,  0.00) rectangle (216.81,101.18);
\end{scope}
\begin{scope}
\path[clip] ( 52.16, 37.94) rectangle (211.31, 95.68);
\definecolor{fillColor}{gray}{0.92}

\path[fill=fillColor] ( 52.16, 37.94) rectangle (211.31, 95.68);
\definecolor{drawColor}{RGB}{255,255,255}

\path[draw=drawColor,line width= 0.3pt,line join=round] ( 52.16, 49.03) --
	(211.31, 49.03);

\path[draw=drawColor,line width= 0.3pt,line join=round] ( 52.16, 65.96) --
	(211.31, 65.96);

\path[draw=drawColor,line width= 0.3pt,line join=round] ( 52.16, 82.89) --
	(211.31, 82.89);

\path[draw=drawColor,line width= 0.3pt,line join=round] ( 83.03, 37.94) --
	( 83.03, 95.68);

\path[draw=drawColor,line width= 0.3pt,line join=round] (130.31, 37.94) --
	(130.31, 95.68);

\path[draw=drawColor,line width= 0.3pt,line join=round] (177.60, 37.94) --
	(177.60, 95.68);

\path[draw=drawColor,line width= 0.6pt,line join=round] ( 52.16, 40.56) --
	(211.31, 40.56);

\path[draw=drawColor,line width= 0.6pt,line join=round] ( 52.16, 57.49) --
	(211.31, 57.49);

\path[draw=drawColor,line width= 0.6pt,line join=round] ( 52.16, 74.43) --
	(211.31, 74.43);

\path[draw=drawColor,line width= 0.6pt,line join=round] ( 52.16, 91.36) --
	(211.31, 91.36);

\path[draw=drawColor,line width= 0.6pt,line join=round] ( 59.38, 37.94) --
	( 59.38, 95.68);

\path[draw=drawColor,line width= 0.6pt,line join=round] (106.67, 37.94) --
	(106.67, 95.68);

\path[draw=drawColor,line width= 0.6pt,line join=round] (153.95, 37.94) --
	(153.95, 95.68);

\path[draw=drawColor,line width= 0.6pt,line join=round] (201.24, 37.94) --
	(201.24, 95.68);
\definecolor{drawColor}{RGB}{51,102,255}

\path[draw=drawColor,line width= 1.1pt,line join=round] ( 59.39, 46.03) --
	( 61.22, 57.58) --
	( 63.06, 84.53) --
	( 64.89, 89.35) --
	( 66.72, 86.89) --
	( 68.55, 91.07) --
	( 70.38, 87.58) --
	( 72.21, 85.52) --
	( 74.04, 81.14) --
	( 75.88, 81.54) --
	( 77.71, 88.17) --
	( 79.54, 82.97) --
	( 81.37, 77.99) --
	( 83.20, 73.61) --
	( 85.03, 76.98) --
	( 86.86, 78.90) --
	( 88.70, 79.73) --
	( 90.53, 84.95) --
	( 92.36, 84.83) --
	( 94.19, 89.06) --
	( 96.02, 88.05) --
	( 97.85, 90.68) --
	( 99.68, 83.57) --
	(101.52, 80.61) --
	(103.35, 81.30) --
	(105.18, 84.43) --
	(107.01, 83.87) --
	(108.84, 87.06) --
	(110.67, 86.17) --
	(112.50, 86.93) --
	(114.34, 87.42) --
	(116.17, 86.92) --
	(118.00, 85.94) --
	(119.83, 82.17) --
	(121.66, 82.71) --
	(123.49, 73.15) --
	(125.32, 79.34) --
	(127.16, 84.42) --
	(128.99, 88.85) --
	(130.82, 89.90) --
	(132.65, 90.36) --
	(134.48, 88.01) --
	(136.31, 88.13) --
	(138.14, 90.24) --
	(139.98, 91.01) --
	(141.81, 91.27) --
	(143.64, 90.17) --
	(145.47, 87.50) --
	(147.30, 84.34) --
	(149.13, 65.87) --
	(150.96, 87.69) --
	(152.80, 79.79) --
	(154.63, 83.18) --
	(156.46, 84.20) --
	(158.29, 76.34) --
	(160.12, 88.99) --
	(161.95, 90.72) --
	(163.78, 90.97) --
	(165.62, 81.45) --
	(167.45, 87.74) --
	(169.28, 86.55) --
	(171.11, 86.63) --
	(172.94, 82.74) --
	(174.77, 85.24) --
	(176.60, 83.82) --
	(178.44, 87.98) --
	(180.27, 90.53) --
	(182.10, 89.63) --
	(183.93, 83.00) --
	(185.76, 79.79) --
	(187.59, 84.23) --
	(189.42, 78.90) --
	(191.26, 72.76) --
	(193.09, 87.99) --
	(194.92, 84.75) --
	(196.75, 89.92) --
	(198.58, 90.73) --
	(200.41, 89.32) --
	(202.24, 90.29) --
	(204.08, 77.18);
\end{scope}
\begin{scope}
\path[clip] (  0.00,  0.00) rectangle (216.81,101.18);
\definecolor{drawColor}{gray}{0.30}

\node[text=drawColor,anchor=base east,inner sep=0pt, outer sep=0pt, scale=  0.88] at ( 47.21, 37.53) {20000};

\node[text=drawColor,anchor=base east,inner sep=0pt, outer sep=0pt, scale=  0.88] at ( 47.21, 54.46) {30000};

\node[text=drawColor,anchor=base east,inner sep=0pt, outer sep=0pt, scale=  0.88] at ( 47.21, 71.40) {40000};

\node[text=drawColor,anchor=base east,inner sep=0pt, outer sep=0pt, scale=  0.88] at ( 47.21, 88.33) {50000};
\end{scope}
\begin{scope}
\path[clip] (  0.00,  0.00) rectangle (216.81,101.18);
\definecolor{drawColor}{gray}{0.20}

\path[draw=drawColor,line width= 0.6pt,line join=round] ( 49.41, 40.56) --
	( 52.16, 40.56);

\path[draw=drawColor,line width= 0.6pt,line join=round] ( 49.41, 57.49) --
	( 52.16, 57.49);

\path[draw=drawColor,line width= 0.6pt,line join=round] ( 49.41, 74.43) --
	( 52.16, 74.43);

\path[draw=drawColor,line width= 0.6pt,line join=round] ( 49.41, 91.36) --
	( 52.16, 91.36);
\end{scope}
\begin{scope}
\path[clip] (  0.00,  0.00) rectangle (216.81,101.18);
\definecolor{drawColor}{gray}{0.20}

\path[draw=drawColor,line width= 0.6pt,line join=round] ( 59.38, 35.19) --
	( 59.38, 37.94);

\path[draw=drawColor,line width= 0.6pt,line join=round] (106.67, 35.19) --
	(106.67, 37.94);

\path[draw=drawColor,line width= 0.6pt,line join=round] (153.95, 35.19) --
	(153.95, 37.94);

\path[draw=drawColor,line width= 0.6pt,line join=round] (201.24, 35.19) --
	(201.24, 37.94);
\end{scope}
\begin{scope}
\path[clip] (  0.00,  0.00) rectangle (216.81,101.18);
\definecolor{drawColor}{gray}{0.30}

\node[text=drawColor,anchor=base,inner sep=0pt, outer sep=0pt, scale=  0.88] at ( 59.38, 26.93) {0};

\node[text=drawColor,anchor=base,inner sep=0pt, outer sep=0pt, scale=  0.88] at (106.67, 26.93) {5000};

\node[text=drawColor,anchor=base,inner sep=0pt, outer sep=0pt, scale=  0.88] at (153.95, 26.93) {10000};

\node[text=drawColor,anchor=base,inner sep=0pt, outer sep=0pt, scale=  0.88] at (201.24, 26.93) {15000};
\end{scope}
\begin{scope}
\path[clip] (  0.00,  0.00) rectangle (216.81,101.18);
\definecolor{drawColor}{RGB}{0,0,0}

\node[text=drawColor,anchor=base,inner sep=0pt, outer sep=0pt, scale=  1.10] at (131.73,  7.64) {Sweep Number};
\end{scope}
\begin{scope}
\path[clip] (  0.00,  0.00) rectangle (216.81,101.18);
\definecolor{drawColor}{RGB}{0,0,0}

\node[text=drawColor,rotate= 90.00,anchor=base,inner sep=0pt, outer sep=0pt, scale=  1.10] at ( 13.08, 66.81) {Point Count};
\end{scope}
\end{tikzpicture}
  \caption[Newer College Valid Count]{Count of valid depth measurements over time for the Newer College Dataset.}
  \label{fig:newer-valid}
\end{figure}
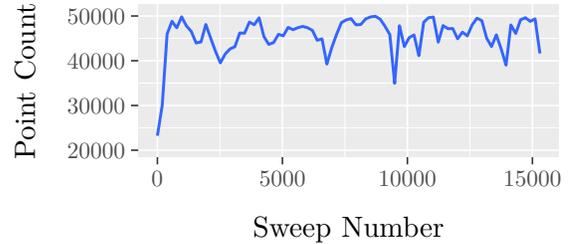

Each incoming panorama of depth and surface normal vectors is registered to the current keyframe using ICP with a point-to-plane error metric \cite{Besl:ICP:1992, Chen:ICP:1992}, Equation~\ref{eq:point2plane}. 

\begin{equation}\label{eq:point2plane}
\mathbf{E}(\mathtt{T}) = \sum_{(s_1,s_2) \in C \wedge \Omega_{d,\theta}(s_1,s_2)} \left\lvert (\mathtt{T} s_2 - s_1) \cdot N(s_1) \right\rvert
\end{equation}

Here, we find a 3D transformation, $\mathtt{T}$, that minimizes the distance between each pair of points drawn from the set of projective correspondences, $C$, when projecting the vector between corresponding points onto the normal vector of the first, $N(s_1)$. A similarity filter, $\Omega$, is applied to the set of correspondences to reject those for which the distance, $d$, between the two points is greater than some threshold, or when the angle, $\theta$, between their associated surface normal vector estimates is too great. This method may be efficiently performed with a GPU \cite{Newcombe:KinectFusion:2011}.

The surface normal estimate for the sweep may be slightly over smoothed by the à trous filter, but this smoothness helps with optimization convergence. The keyframe surface normal estimates are updated by each sweep, which restores some of the detail lost to smoothing. A critical factor in the robustness of the mapping system is how many points are incorporated into this optimization. Figure~\ref{fig:newer-valid} shows the number of valid points over time for a typical experiment, where our notion of validity is defined as a point deriving from a valid range estimate (the sensor will not return a measurement for every pixel in every sweep), and an intensity above a small threshold (this rejects points whose range estimate is based on very little received light). In the example shown in the figure, the sensor starts out near a wall in a tunnel, but once it emerges into the open, 40 to 50 thousand points are used at the 10Hz sweep rate.

\subsection{Keyframe Update}
The optimized transformation provides a basis for a final set of projective correspondences, in which each keyframe pixel is projected onto the sweep depth image. Just as with the ICP optimization, a similarity filter is used to determine if the new sweep data should be averaged in to the keyframe pixel. A weighted average is used when updating the keyframe: each keyframe pixel tracks how many samples have been incorporated into it, and that number -- saturated at 10 in our system -- is used to determine the weightings of the averaging procedure. The simplicity of this averaging scheme is offset by the use of multiple keyframes in the map representation when considered holistically. Moving objects can strain the balance of smoothing and responsiveness within a single keyframe, but considering multiple keyframes reconstructs a latent multi-hypothesis model of 3D occupancy which is an effective tool for filtering moving objects from the map. This mutual consistency check between keyframes is performed by UPSLAM at runtime whenever switching keyframes in order to reduce the impact of moving objects.

Once updated, a decision is made as to whether we should continue using the current keyframe. If the registration quality -- defined as the ratio of ICP registration inliers to valid depth measurements in the sweep -- has dropped below a configurable threshold \footnote{This configuration parameter was allowed to vary between experiments.}, a new keyframe is sought. The pose graph is first consulted to identify the nearest keyframes to the current pose estimate to detect potential loop closures.

Since these loop closure registrations are likely to occur under the influence of significant drift, a grid search is undertaken in which several hundred transformations are evaluated with low-resolution projective alignment checks. The best candidate found by this search is promoted to a full-resolution ICP registration. If the registration quality is higher than the threshold for creating a new keyframe, the pose graph is updated with an edge between the current keyframe and the neighbor, and then that neighbor becomes the new current keyframe. Otherwise, if that ratio is higher than 75\% of the new keyframe inlier ratio, the pose graph is updated with the transformation computed from the registration, but the current keyframe is not updated. We refer to these as \textit{strong} and \textit{weak} loop closures: both varieties improve the global accuracy of the map, but strong loop closures additionally reduce the number of keyframes created when a location is re-visited.

The pose graph structure is reflected in GTSAM \cite{Dellaert2012FactorGA}, which is used to globally optimize inter-keyframe transformations on the CPU. Finally, a complementary filter \cite{Valenti:IMU:2015} on the IMU data is used to maintain a smoothed estimate of Earth's gravity vector which is attached to each keyframe as it is added to GTSAM. The observations of this shared phenomenon help minimize the effects of errors in pitch, which are relatively common given the limited vertical field of view of the sensor.

\section{Qualitative Breadth}
\subsection{University - Indoor}
\begin{figure}
\centering
\subcaptionbox*{}{\includegraphics[width=0.3\columnwidth]{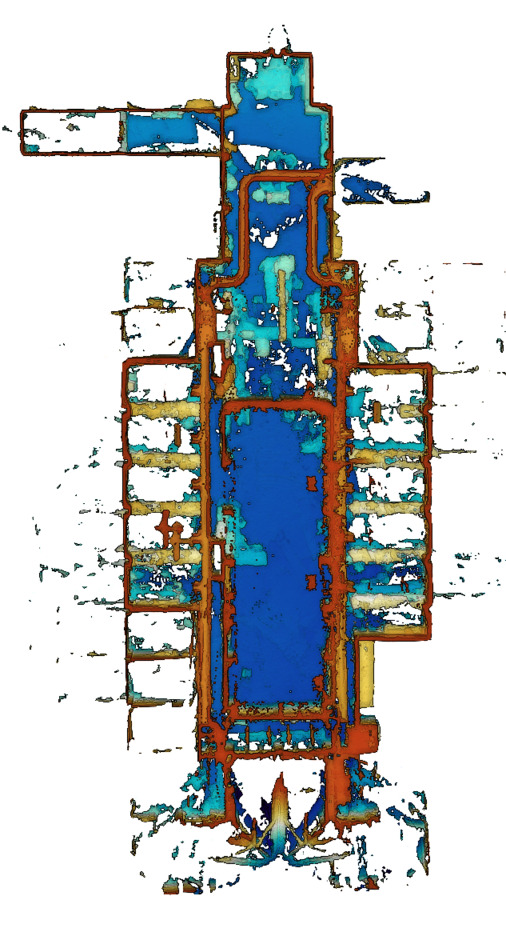}}
\subcaptionbox*{}{\includegraphics[width=0.3\columnwidth]{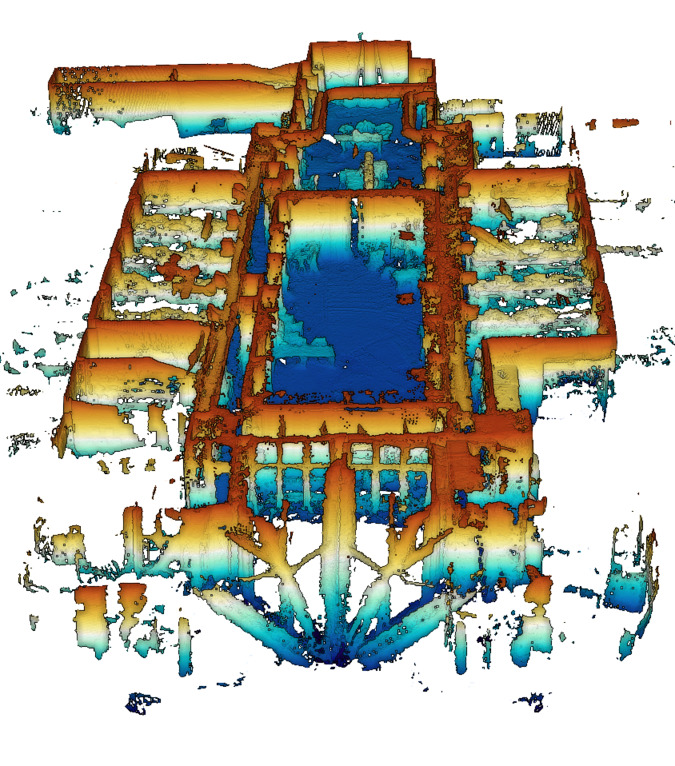}}
\caption[PERCH Lab Interior]{A \SI{145}{m} trajectory through a cluttered lab space. The ceiling was removed for visualization clarity.}
\label{fig:perch}
\end{figure}

A first test of the mapping software is a \SI{145}{m} walk around a lab space, moving at speed of \SIrange[range-phrase = --, range-units = single]{1}{2}{m/s}. The map shown in Figure~\ref{fig:perch} was captured with 50 keyframe images occupying \SI{80}{MiB} on disk. While the motion here is benign, the abundance of glass around the angular section at the bottom of the figure, along with a few narrow passages, provides some challenge due to the sensor's inability to reliably detect glass or walls closer than around \SI{50}{cm}.

\subsection{University - Outdoor}

\begin{figure}
\centering
\subcaptionbox{\label{fig:penno-front}}{\includegraphics[width=0.9\columnwidth]{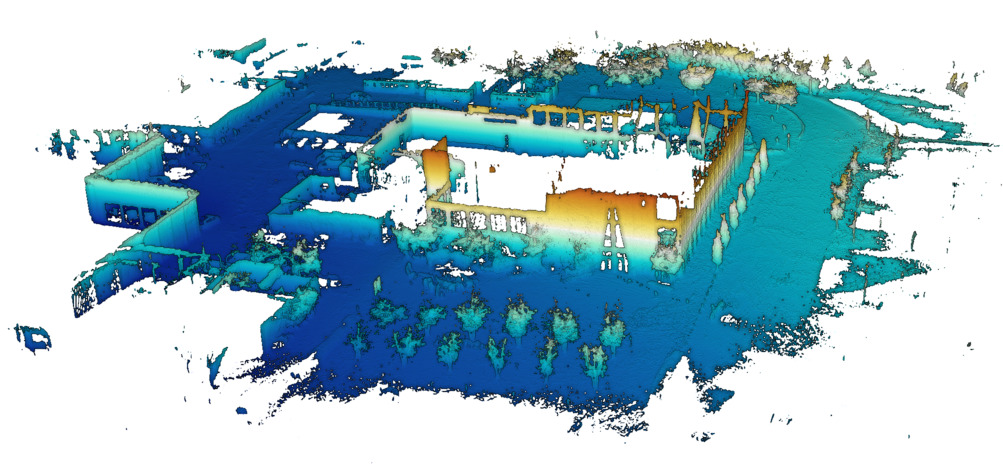}}
\subcaptionbox{\label{fig:penno-side}}{\includegraphics[width=0.9\columnwidth]{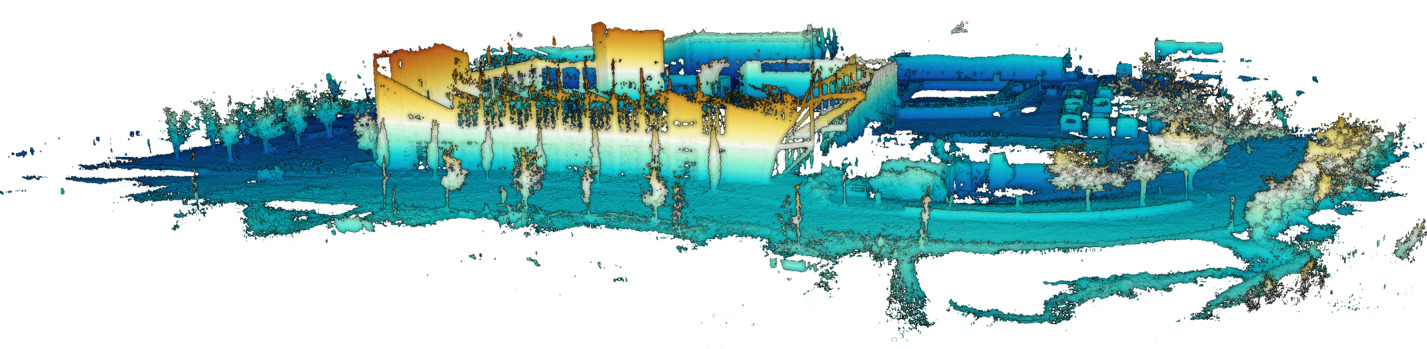}}
\caption[Pennovation Exterior]{A running loop around the Pennovation campus with the sensor carried overhead.}
\label{fig:pennovation}
\end{figure}

The sensor was then taken outside, carried overhead, and run along a \SI{375}{m} loop around a cluster of buildings on a University campus at \SIrange[range-phrase = --, range-units = single]{2}{3}{m/s}. The resulting point cloud, shown in Figure~\ref{fig:pennovation}, benefits from the single loop closure formed when the sensor came within approximately \SI{10}{m} of its starting location near the trees in the lower part of Figure~\ref{fig:penno-front}. The running pace was chosen to stress non-planar motion with high amplitude rotations due to shaking and bouncing.

\subsection{Farm}

\begin{figure}[!ht]
\centering
\subcaptionbox{\label{fig:farm-jackal}}{\includegraphics[width=0.45\columnwidth]{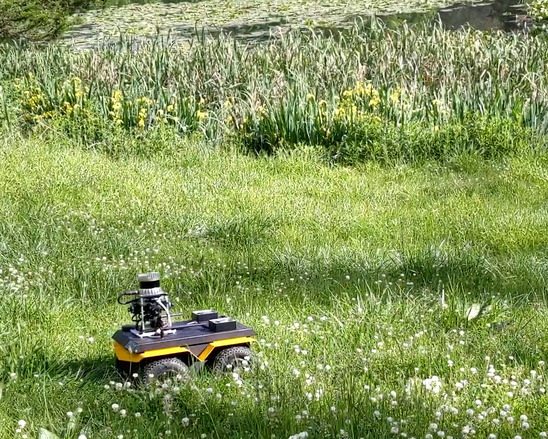}}
\subcaptionbox{\label{fig:farm-earth}}{\includegraphics[width=0.45\columnwidth]{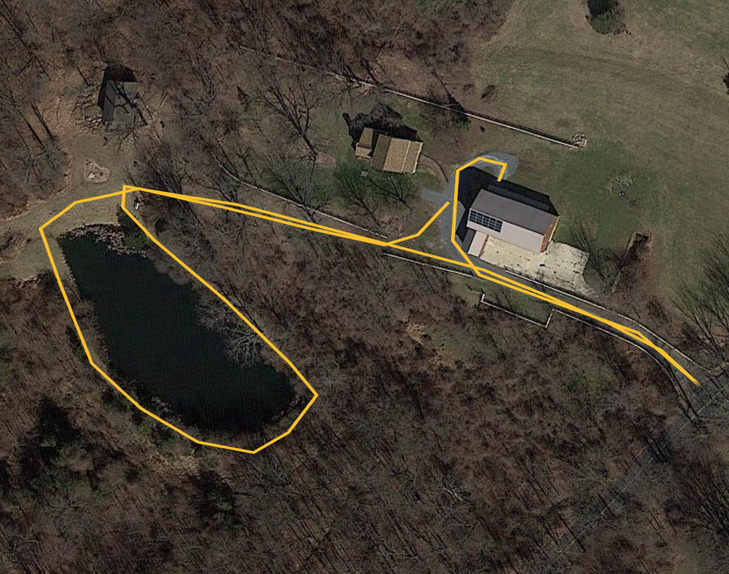}}
\subcaptionbox{\label{fig:farm-points}}{\includegraphics[width=0.9\columnwidth]{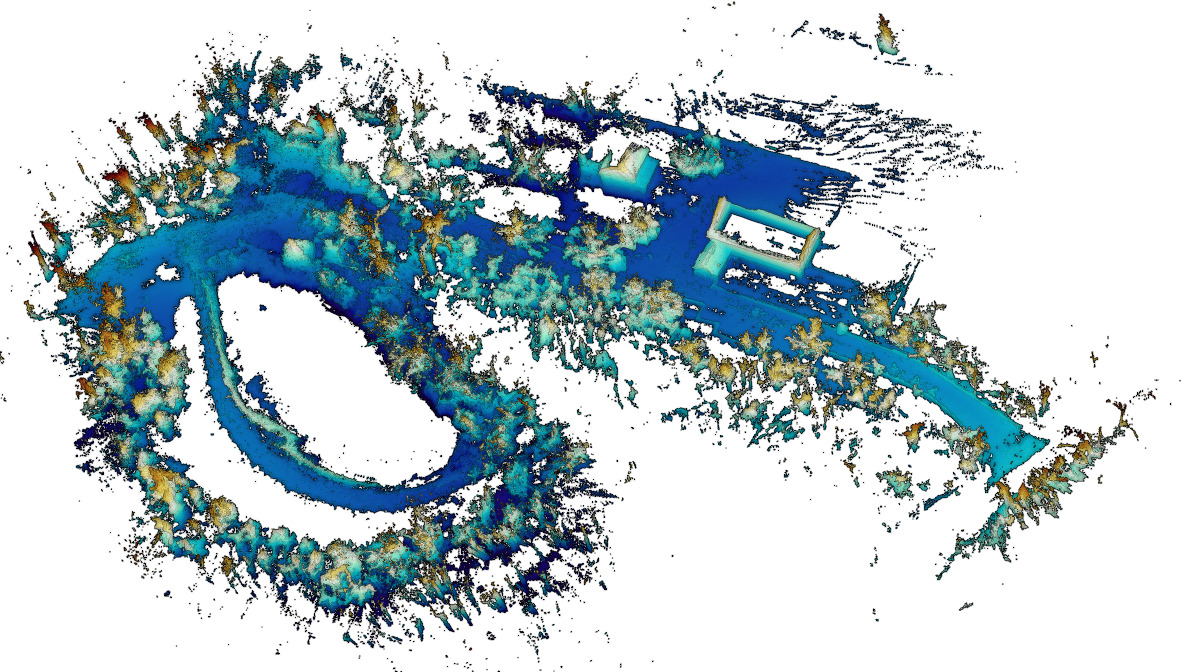}}
\subcaptionbox{\label{fig:farm-points-fence}}{\includegraphics[width=0.75\columnwidth]{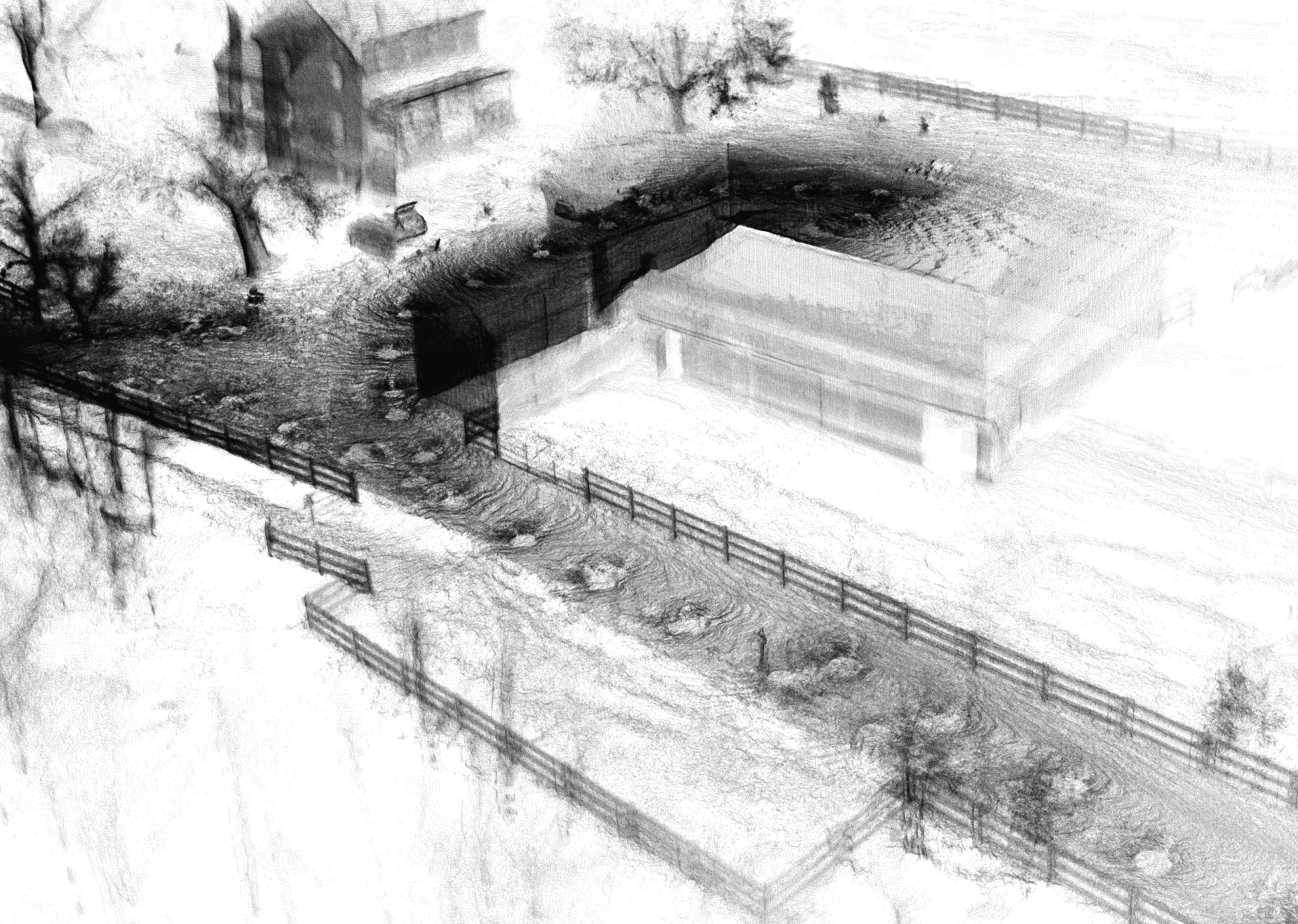}}
\caption[Offroad Driving at Farm]{The Miller family farm in central Pennsylvania \subref{fig:farm-earth}. The 731m on- and off-road trajectory around pond is captured with geometric detail including individual rails on gates and fences. \subref{fig:farm-points-fence}.}
\end{figure}

A Clearpath Robotics Jackal, Figure~\ref{fig:farm-jackal}, was driven along a \SI{731}{m} route along a driveway, dirt trail, and grass on a farm, Figure~\ref{fig:farm-earth}. This excursion presented several challenges due to the rough terrain and varied surroundings. Tall vegetation waved in the breeze, while the pond itself offered minimal range measurements. A steep drop to the side of the raised trail around the pond prevented the sensor from seeing any nearby ground (the left of Figure~\ref{fig:farm-points}). The path taken involved mud, logs, and vegetation that the vehicle could not always pass directly through, so the data collection included intermittent occlusion of the sensor, along with sharp accelerations to get the vehicle un-stuck.

\subsection{Underground}
\begin{figure}[!ht]
\centering
\subcaptionbox{\label{fig:vision60}}{\includegraphics[width=0.45\columnwidth]{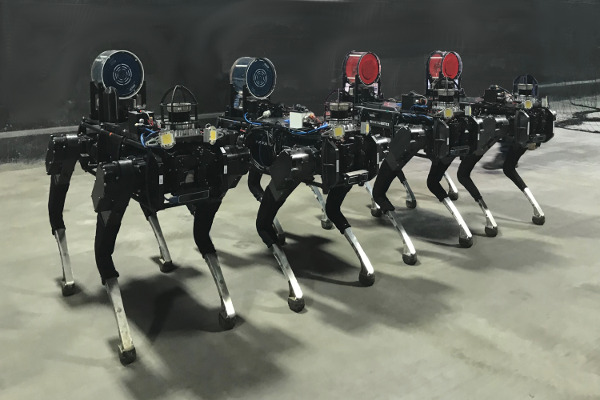}}
\subcaptionbox{\label{fig:no9}}{\includegraphics[width=0.25\columnwidth]{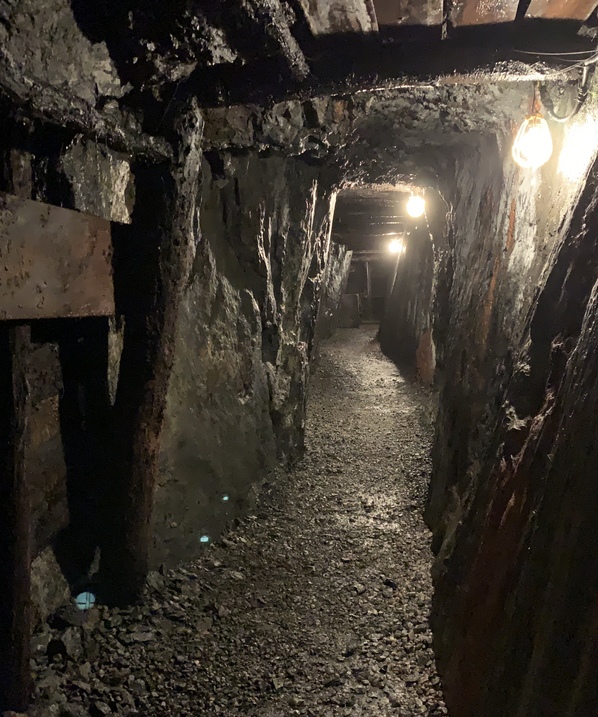}}
\subcaptionbox{\label{fig:mine-points}}{\includegraphics[width=0.9\columnwidth]{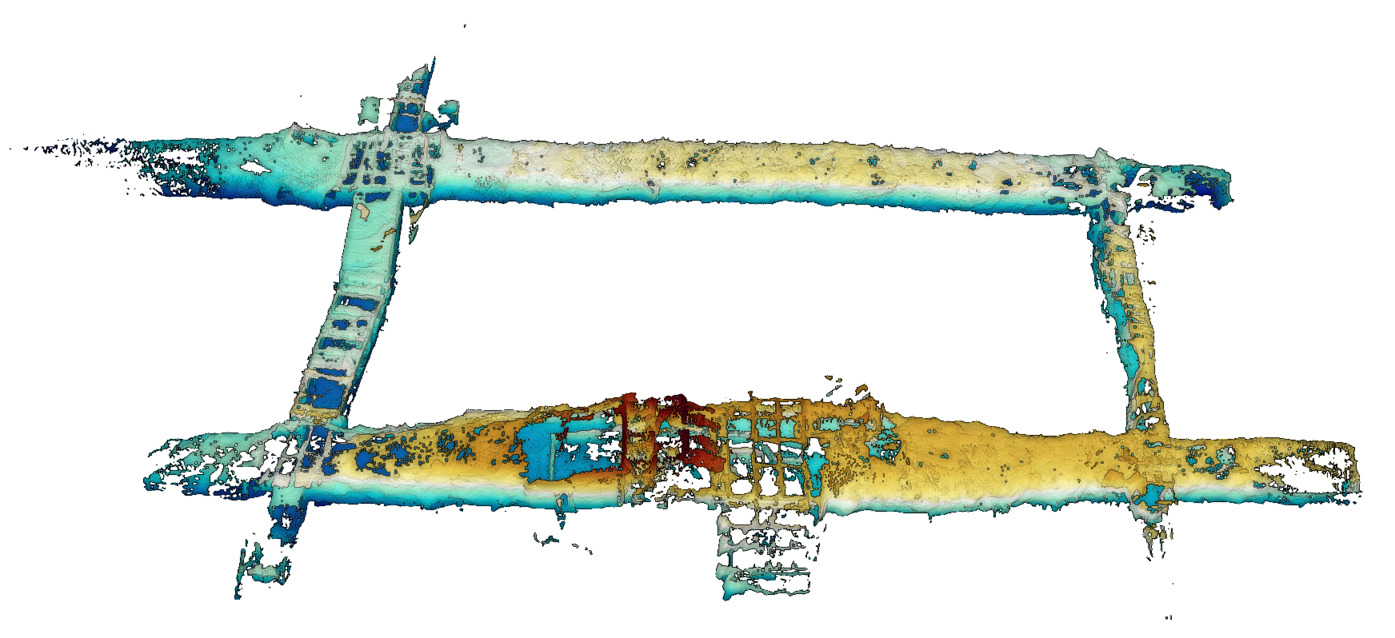}}
\caption[No. 9 Coal Mine]{A quadrupedal robot \subref{fig:vision60} was walked around a coal mine \subref{fig:no9} along a 220m trajectory. The resulting map \subref{fig:mine-points} was captured with 44 images.}
\label{fig:tourist}
\end{figure}
A Ghost Robotics Vision 60 quadruped robot was walked along a 220m trajectory in an old coal mine, capturing the map shown in Figure~\ref{fig:tourist} with just 44 panoramic images. This experiment provided a substantially different motion model than those previously described as the wet gravel floor of the mine meant that the robot had to continually work to maintain its balance. Dripping water, mist, and dust contribute a substantial amount of noise to individual lidar sweeps. Additionally, the tunnels visited ranged from wide access to the main elevator, bottom-center of Figure~\ref{fig:mine-points}, to a narrow, angular tunnel approximately \SI{1}{m} across, Figure~\ref{fig:no9}, visible on the right side of the point cloud. The map is constructed despite the presence of short bouts of uncertain footing for the robot due to changes in wet, sloppy floor material, and a trajectory through the narrowest of walkways that involved the robot bumping from wall to wall.

\subsection{Aerial}

\begin{figure}[!ht]
\centering
\subcaptionbox{\label{fig:quadrotor-forest}}{\includegraphics[width=0.35\columnwidth]{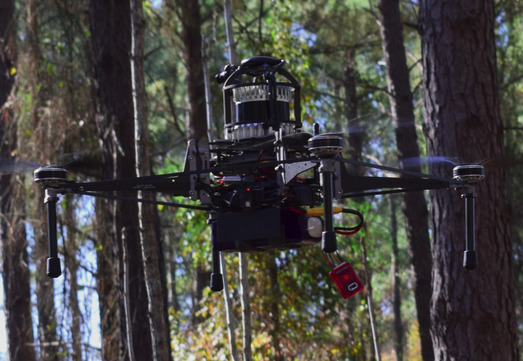}}
\\
\subcaptionbox{\label{fig:wharton-eyedome}}{\includegraphics[width=0.45\columnwidth]{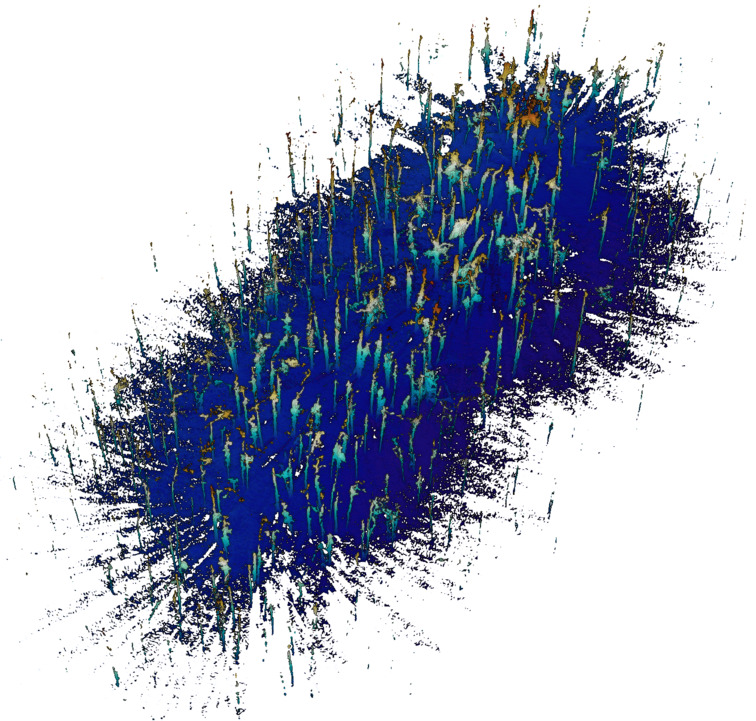}}
\subcaptionbox{\label{fig:wharton-points-trans}}{\includegraphics[width=0.45\columnwidth]{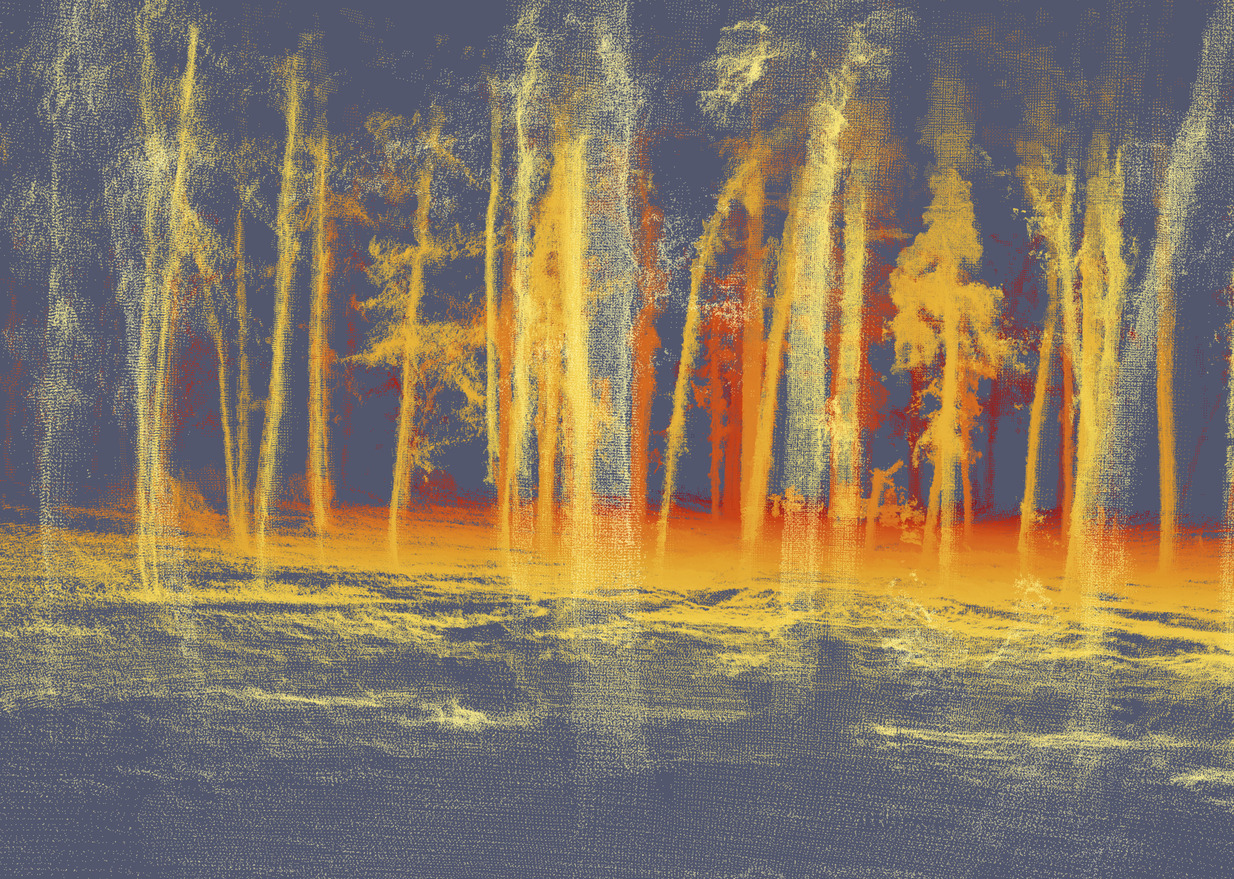}}
\caption[Wharton State Forest]{A quadrotor was flown along a \SI{117}{m} trajectory in a dense forest. The entire environment \subref{fig:wharton-eyedome} is devoid of manmade structure, but the trunks and branches of the many trees are cleanly captured in the point cloud \subref{fig:wharton-points-trans}.}
\label{fig:wharton}
\end{figure}

The Wharton State Forest in New Jersey is an expansive wilderness with dense tree cover. Flying a quadrotor\footnote{Data courtesy of Treeswift.}, Figure~\ref{fig:quadrotor-forest}, in this environment is challenging for many reasons, among which is the challenge of building a map in a region with no large planar surfaces. Occlusion edges, too, can be difficult to work with as they are virtually all tree trunks without a geometrically simple background. One moment a sight line passing the edge of a tree trunk is too long for the lidar to reliably measure, but the next moment, from a slightly different angle, the same tree trunk partially occludes another nearby tree. UPSLAM's use of dense surface measurements extracts as much as possible from the sensor, and it is able to track its motion along a \SI{117}{m} trajectory while producing a map populated with crisply defined trees, Figure~\ref{fig:wharton}. This map is captured in 30 keyframe images occupying \SI{40}{MiB} on disk, sufficient to produce the point cloud shown in Figure~\ref{fig:wharton-points-trans}. The union of panoramic images employed by UPSLAM is able to efficiently capture this unstructured 3D geometry.

\subsection{Automotive}
\begin{figure}
\centering
\subcaptionbox{\label{fig:morgtown-gearth}}{\includegraphics[width=0.7\columnwidth]{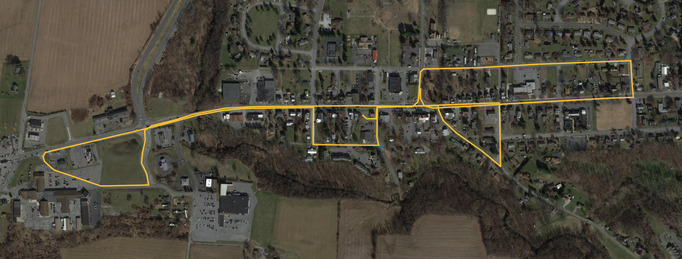}}
\subcaptionbox{\label{fig:morgtown-video}}{\includegraphics[width=0.7\columnwidth]{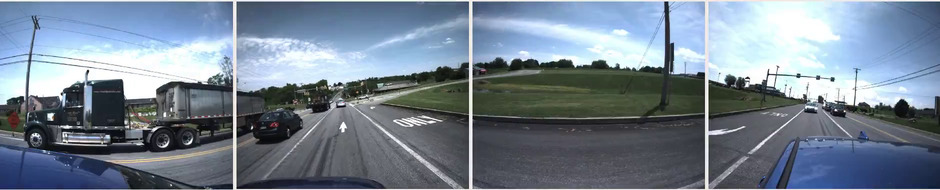}}
\subcaptionbox{\label{fig:morgtown-points-above}}{\includegraphics[width=0.7\columnwidth]{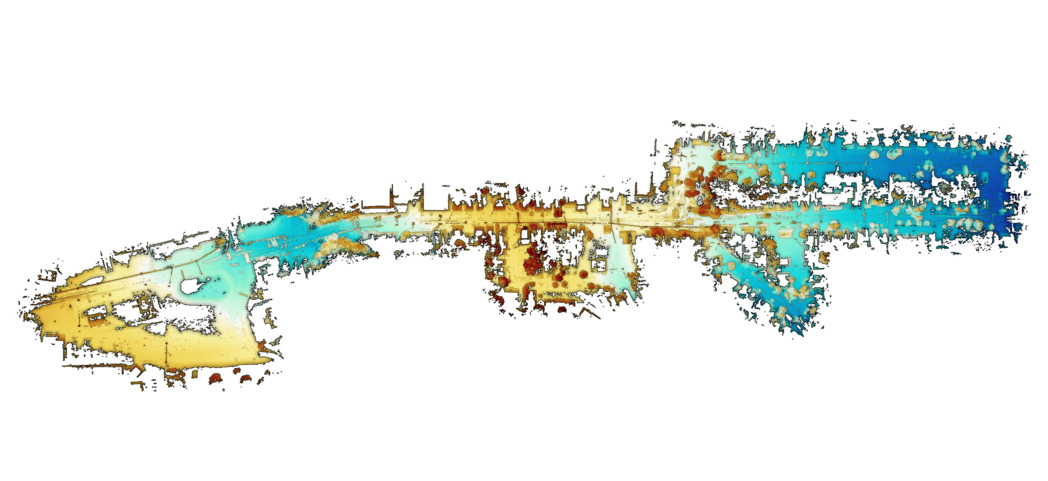}}
\caption[Morgantown, PA, satellite view]{A 4.4km route driven through Morgantown, Pennsylvania, highlighted in Google Earth \subref{fig:morgtown-gearth}, seen from street level \subref{fig:morgtown-video}, and viewed as a point cloud produced by UPSLAM \subref{fig:morgtown-points-above}.}
\label{fig:morgtown-above}
\end{figure}

While each of the previous experiments subjected the system to different motion models, the fastest the sensor was moved was about \SI{3}{m/s} when carried by a running human. The Ouster OS1-64 was scanning at \SI{10}{Hz} for all of these data collections, meaning that a single sweep would see the sensor moved by up to \SI{30}{cm}. This section presents a dramatically different scenario: the sensor was mounted atop an automobile and driven \SI{4.4}{km} through a town at up to \SI{50}{km/h} (31 miles per hour), Figure~\ref{fig:morgtown-above}.

\begin{figure}
\centering
\subcaptionbox{\label{fig:morgtown-angle}}{\includegraphics[width=0.45\columnwidth]{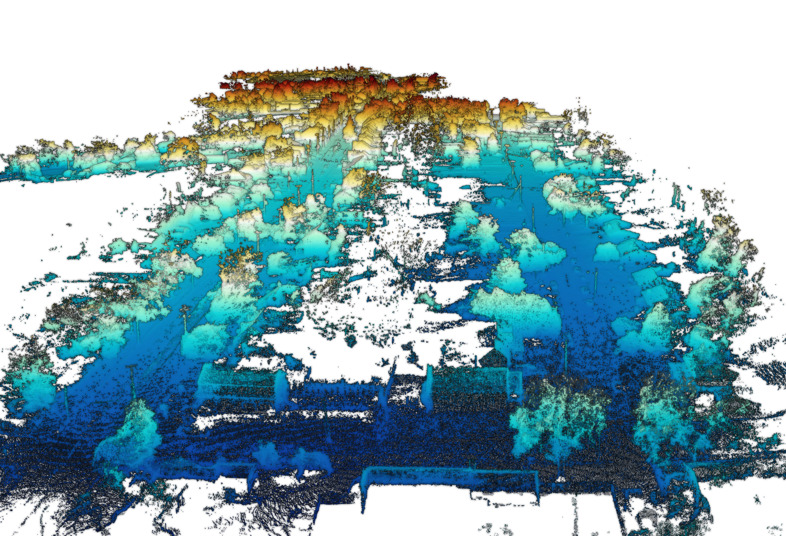}}
\subcaptionbox{\label{fig:morgtown-angle-trans}}{\includegraphics[width=0.45\columnwidth]{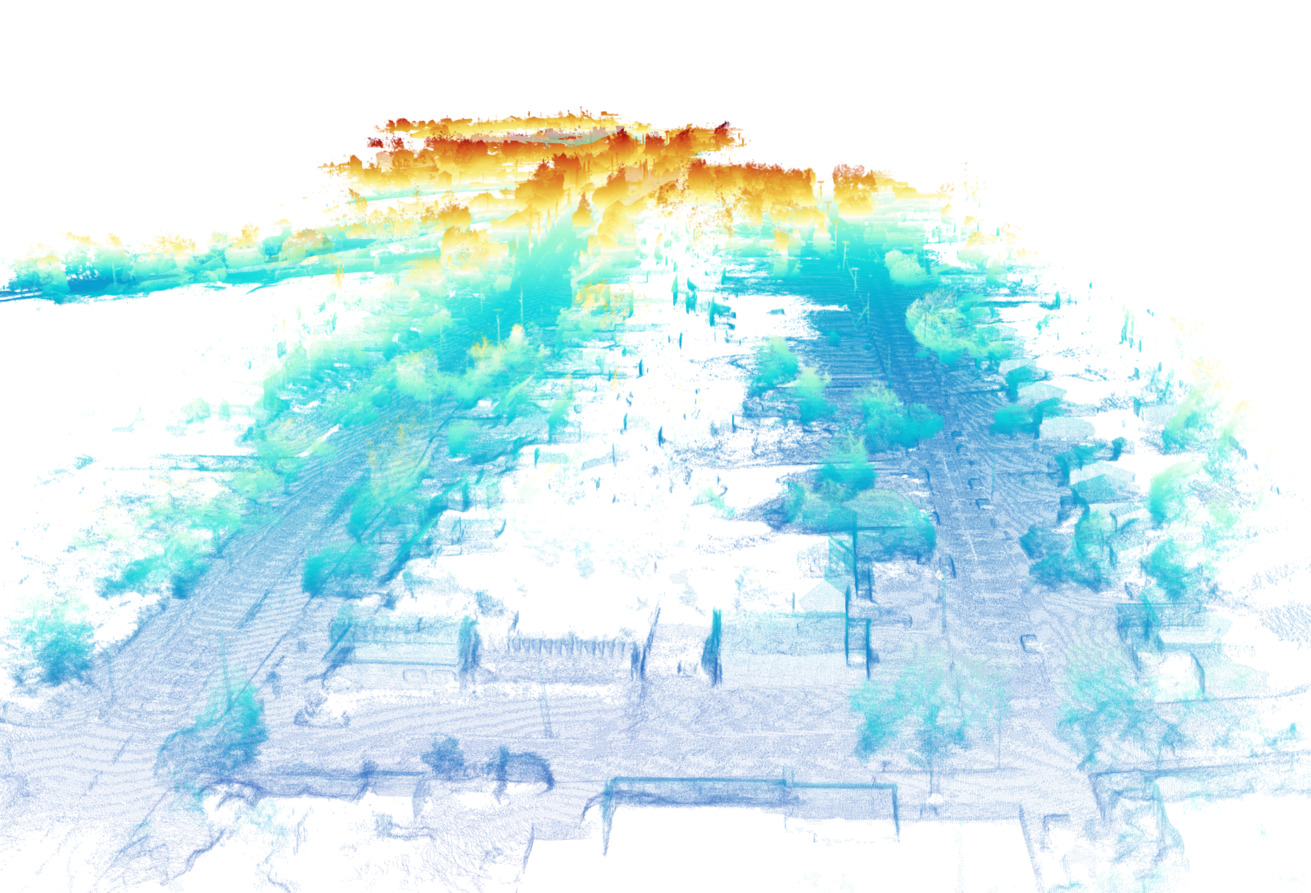}}
\subcaptionbox{\label{fig:morgtown-sdf1}}{\includegraphics[width=0.9\columnwidth]{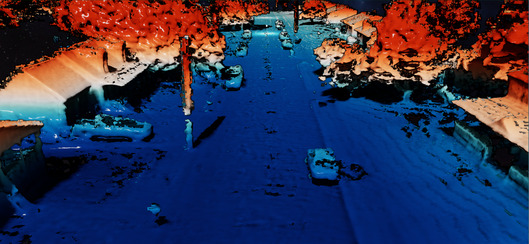}}
\caption[Morgantown, PA, point clouds]{A lower view of the Morgantown dataset shows some of the structural detail picked up by the single sensor moving at road speeds.}
\label{fig:morgtown-points}
\end{figure}

This fast-moving data set\footnote{A second configuration parameter was used for this experiment to enable a zero acceleration motion prior, as opposed to the default zero velocity prior used in all other experiments which involve more sudden reversals of velocity.} prompted 70 loop closures, and produced 1045 keyframe images occupying 1220MiB on disk. Map construction involved 20GiB of raw, uncompressed keyframe data at runtime, however UPSLAM is able to stream keyframes between disk and GPU memory as needed, allowing us to cap the resident set at 3GiB of GPU memory. Because the sensor was moving so quickly -- up to \SI{1.4}{m} per sweep -- there is far less redundancy in the collected lidar point cloud, limiting opportunities for compression, moving object filtering, or even surface smoothing. However, considerable structural detail is still apparent despite the presence of other vehicles on the road, Figure~\ref{fig:morgtown-points}. Shown are renderings of the road at the top right of the trajectory shown in Figure~\ref{fig:morgtown-above} viewed from the right edge looking left. An opaque point cloud rendering highlights depth, Figure~\ref{fig:morgtown-angle}, while the same point cloud with lowered opacity demonstrates fine geometric detail, Figure~\ref{fig:morgtown-angle-trans}. Rendering the scene with a real time ray marched lighting model, Figure~\ref{fig:morgtown-sdf1}, provides evidence that the map truly captures the three dimensional scene with surface shading, shadows, and occlusion.

\section{Quantitative Depth}
\begin{figure}
\centering
\subcaptionbox{\label{fig:newer-overhead}}{\includegraphics[width=0.66\columnwidth]{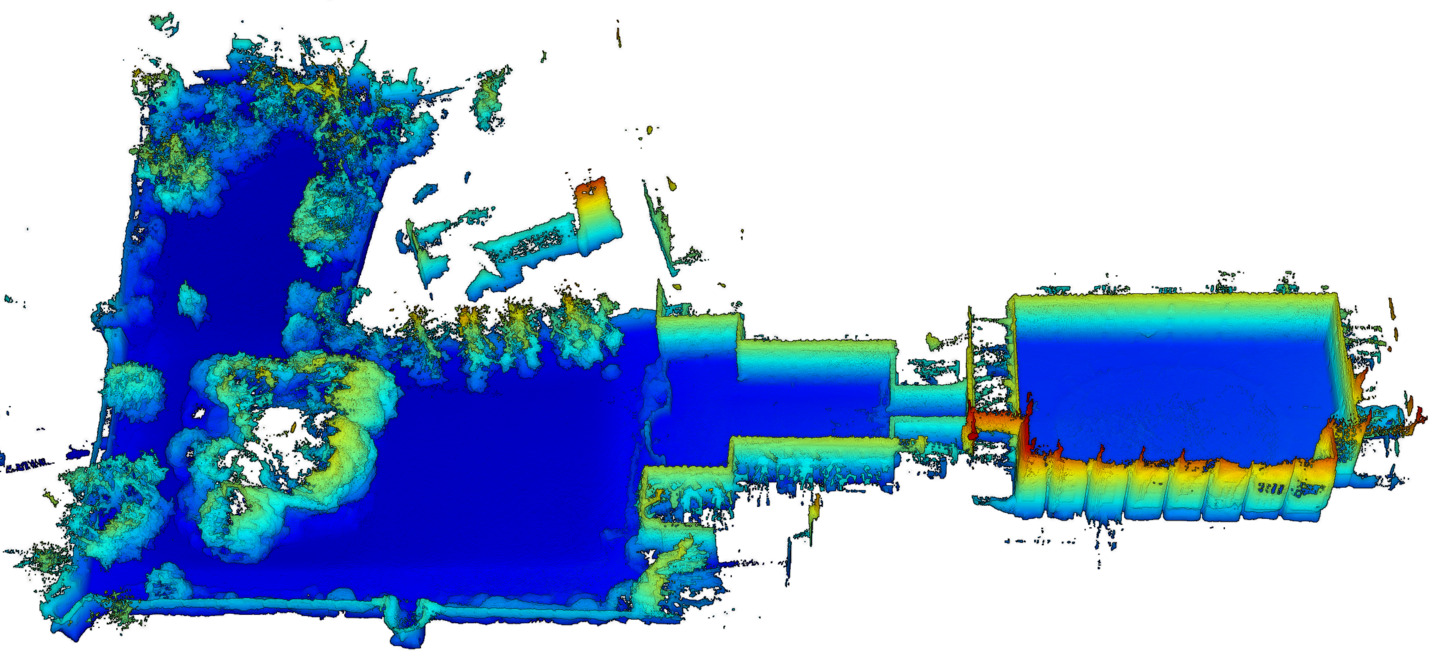}}
\subcaptionbox{\label{fig:newer-sdf}}{\includegraphics[width=0.745\columnwidth]{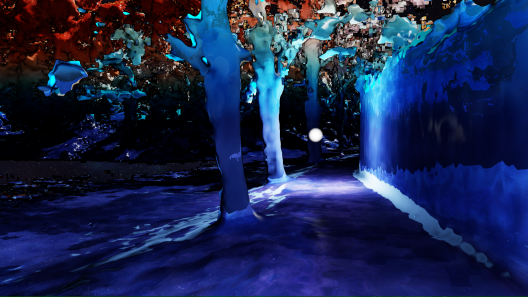}}
\caption{An overhead view of a point cloud representing New College, Oxford University \subref{fig:newer-overhead} provides a sense of scale, while dropping to ground level between trees and a wall in the park section, \subref{fig:newer-sdf}, shows geometry density.}
\label{fig:newer-points}
\end{figure}

The Newer College Dataset \cite{ramezani2020newer} provides an opportunity to quantify trajectory accuracy. Point clouds captured by an Ouster OS1-64 were registered against a point cloud collected by a tripod-mounted, survey grade Leica BLK360 scanner. A sense of the space may be found in the point cloud rendering shown in Figure~\ref{fig:newer-overhead}, showing the quad on the lower right, the midsection, and then the park on the left. The sensor was carried on a circuitous route making two full laps around the college grounds \footnote{This is the \texttt{short\_experiment} of the dataset.}. 

\subsection{Trajectory Accuracy}
\begin{figure}
  \centering
  \vspace*{-12mm}
  \input{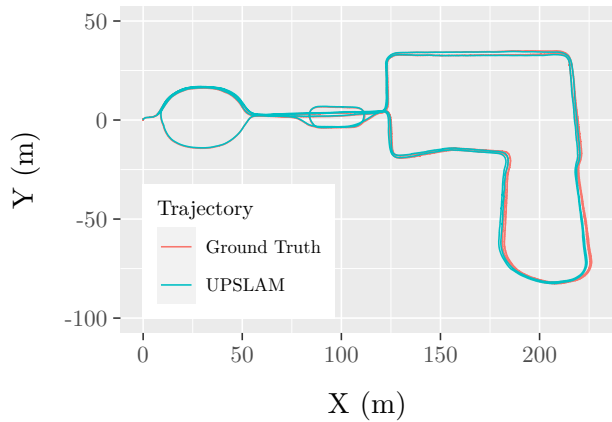}
  \vspace*{-12mm}
  \caption[Newer College trajectories]{Comparison of ground truth and UPSLAM trajectories for the Newer College Dataset.}
  \label{fig:newer-traj}
\end{figure}

\begin{figure}
  \centering
\begin{tikzpicture}[x=1pt,y=1pt]
\definecolor{fillColor}{RGB}{255,255,255}
\path[use as bounding box,fill=fillColor,fill opacity=0.00] (0,0) rectangle (216.81, 86.72);
\begin{scope}
\path[clip] (  0.00,  0.00) rectangle (216.81, 86.72);
\definecolor{drawColor}{RGB}{255,255,255}
\definecolor{fillColor}{RGB}{255,255,255}

\path[draw=drawColor,line width= 0.6pt,line join=round,line cap=round,fill=fillColor] (  0.00,  0.00) rectangle (216.81, 86.72);
\end{scope}
\begin{scope}
\path[clip] ( 41.41, 30.69) rectangle (211.31, 81.22);
\definecolor{fillColor}{gray}{0.92}

\path[fill=fillColor] ( 41.41, 30.69) rectangle (211.31, 81.22);
\definecolor{drawColor}{RGB}{255,255,255}

\path[draw=drawColor,line width= 0.3pt,line join=round] ( 41.41, 36.62) --
	(211.31, 36.62);

\path[draw=drawColor,line width= 0.3pt,line join=round] ( 41.41, 51.09) --
	(211.31, 51.09);

\path[draw=drawColor,line width= 0.3pt,line join=round] ( 41.41, 65.56) --
	(211.31, 65.56);

\path[draw=drawColor,line width= 0.3pt,line join=round] ( 41.41, 80.03) --
	(211.31, 80.03);

\path[draw=drawColor,line width= 0.3pt,line join=round] ( 74.57, 30.69) --
	( 74.57, 81.22);

\path[draw=drawColor,line width= 0.3pt,line join=round] (125.45, 30.69) --
	(125.45, 81.22);

\path[draw=drawColor,line width= 0.3pt,line join=round] (176.33, 30.69) --
	(176.33, 81.22);

\path[draw=drawColor,line width= 0.6pt,line join=round] ( 41.41, 43.86) --
	(211.31, 43.86);

\path[draw=drawColor,line width= 0.6pt,line join=round] ( 41.41, 58.33) --
	(211.31, 58.33);

\path[draw=drawColor,line width= 0.6pt,line join=round] ( 41.41, 72.80) --
	(211.31, 72.80);

\path[draw=drawColor,line width= 0.6pt,line join=round] ( 49.13, 30.69) --
	( 49.13, 81.22);

\path[draw=drawColor,line width= 0.6pt,line join=round] (100.01, 30.69) --
	(100.01, 81.22);

\path[draw=drawColor,line width= 0.6pt,line join=round] (150.89, 30.69) --
	(150.89, 81.22);

\path[draw=drawColor,line width= 0.6pt,line join=round] (201.77, 30.69) --
	(201.77, 81.22);
\definecolor{drawColor}{RGB}{51,102,255}

\path[draw=drawColor,line width= 1.1pt,line join=round] ( 49.13, 59.06) --
	( 51.08, 59.40) --
	( 53.04, 59.24) --
	( 54.99, 57.10) --
	( 56.95, 50.80) --
	( 58.90, 46.70) --
	( 60.86, 41.49) --
	( 62.82, 38.79) --
	( 64.77, 38.01) --
	( 66.73, 39.67) --
	( 68.68, 51.49) --
	( 70.64, 39.70) --
	( 72.59, 39.31) --
	( 74.55, 38.26) --
	( 76.50, 36.10) --
	( 78.46, 34.93) --
	( 80.41, 32.98) --
	( 82.37, 33.78) --
	( 84.32, 40.30) --
	( 86.28, 42.85) --
	( 88.23, 47.21) --
	( 90.19, 42.29) --
	( 92.14, 38.36) --
	( 94.10, 39.90) --
	( 96.05, 50.96) --
	( 98.01, 62.18) --
	( 99.96, 68.38) --
	(101.92, 68.80) --
	(103.87, 65.09) --
	(105.83, 61.75) --
	(107.78, 54.10) --
	(109.74, 50.67) --
	(111.69, 39.35) --
	(113.65, 43.10) --
	(115.60, 50.81) --
	(117.56, 42.44) --
	(119.51, 41.86) --
	(121.47, 37.58) --
	(123.43, 47.56) --
	(125.38, 52.27) --
	(127.34, 52.28) --
	(129.29, 55.29) --
	(131.25, 55.62) --
	(133.20, 51.67) --
	(135.16, 53.35) --
	(137.11, 51.83) --
	(139.07, 44.87) --
	(141.02, 38.44) --
	(142.98, 36.92) --
	(144.93, 39.35) --
	(146.89, 46.01) --
	(148.84, 46.08) --
	(150.80, 48.10) --
	(152.75, 49.34) --
	(154.71, 46.87) --
	(156.66, 47.50) --
	(158.62, 58.13) --
	(160.57, 68.87) --
	(162.53, 69.49) --
	(164.48, 64.25) --
	(166.44, 78.93) --
	(168.39, 78.30) --
	(170.35, 65.08) --
	(172.30, 54.68) --
	(174.26, 44.24) --
	(176.21, 40.51) --
	(178.17, 41.58) --
	(180.13, 50.21) --
	(182.08, 49.43) --
	(184.04, 46.70) --
	(185.99, 39.12) --
	(187.95, 39.01) --
	(189.90, 40.50) --
	(191.86, 41.39) --
	(193.81, 48.37) --
	(195.77, 52.84) --
	(197.72, 51.93) --
	(199.68, 55.65) --
	(201.63, 56.69) --
	(203.59, 57.06);
\end{scope}
\begin{scope}
\path[clip] (  0.00,  0.00) rectangle (216.81, 86.72);
\definecolor{drawColor}{gray}{0.30}

\node[text=drawColor,anchor=base east,inner sep=0pt, outer sep=0pt, scale=  0.88] at ( 36.46, 40.83) {0.5};

\node[text=drawColor,anchor=base east,inner sep=0pt, outer sep=0pt, scale=  0.88] at ( 36.46, 55.30) {1.0};

\node[text=drawColor,anchor=base east,inner sep=0pt, outer sep=0pt, scale=  0.88] at ( 36.46, 69.77) {1.5};
\end{scope}
\begin{scope}
\path[clip] (  0.00,  0.00) rectangle (216.81, 86.72);
\definecolor{drawColor}{gray}{0.20}

\path[draw=drawColor,line width= 0.6pt,line join=round] ( 38.66, 43.86) --
	( 41.41, 43.86);

\path[draw=drawColor,line width= 0.6pt,line join=round] ( 38.66, 58.33) --
	( 41.41, 58.33);

\path[draw=drawColor,line width= 0.6pt,line join=round] ( 38.66, 72.80) --
	( 41.41, 72.80);
\end{scope}
\begin{scope}
\path[clip] (  0.00,  0.00) rectangle (216.81, 86.72);
\definecolor{drawColor}{gray}{0.20}

\path[draw=drawColor,line width= 0.6pt,line join=round] ( 49.13, 27.94) --
	( 49.13, 30.69);

\path[draw=drawColor,line width= 0.6pt,line join=round] (100.01, 27.94) --
	(100.01, 30.69);

\path[draw=drawColor,line width= 0.6pt,line join=round] (150.89, 27.94) --
	(150.89, 30.69);

\path[draw=drawColor,line width= 0.6pt,line join=round] (201.77, 27.94) --
	(201.77, 30.69);
\end{scope}
\begin{scope}
\path[clip] (  0.00,  0.00) rectangle (216.81, 86.72);
\definecolor{drawColor}{gray}{0.30}

\node[text=drawColor,anchor=base,inner sep=0pt, outer sep=0pt, scale=  0.88] at ( 49.13, 19.68) {0};

\node[text=drawColor,anchor=base,inner sep=0pt, outer sep=0pt, scale=  0.88] at (100.01, 19.68) {500};

\node[text=drawColor,anchor=base,inner sep=0pt, outer sep=0pt, scale=  0.88] at (150.89, 19.68) {1000};

\node[text=drawColor,anchor=base,inner sep=0pt, outer sep=0pt, scale=  0.88] at (201.77, 19.68) {1500};
\end{scope}
\begin{scope}
\path[clip] (  0.00,  0.00) rectangle (216.81, 86.72);
\definecolor{drawColor}{RGB}{0,0,0}

\node[text=drawColor,anchor=base,inner sep=0pt, outer sep=0pt, scale=  1.10] at (126.36,  7.64) {time (s)};
\end{scope}
\begin{scope}
\path[clip] (  0.00,  0.00) rectangle (216.81, 86.72);
\definecolor{drawColor}{RGB}{0,0,0}

\node[text=drawColor,rotate= 90.00,anchor=base,inner sep=0pt, outer sep=0pt, scale=  1.10] at ( 13.08, 55.95) {error (m)};
\end{scope}
\end{tikzpicture}
  \caption[Newer College trajectory error]{Error over time between UPSLAM and ground truth for the Newer College Dataset.}
  \label{fig:newer-error}
\end{figure}
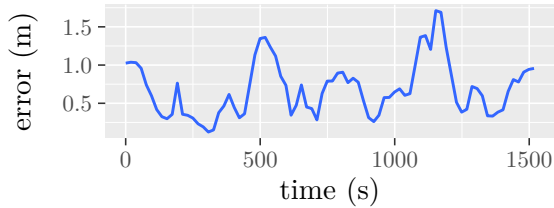

Using time stamps to align the UPSLAM and ground truth trajectories shows broad agreement, Figure~\ref{fig:newer-traj}, with most of the discrepancy introduced around a particular corner of the park area visible at the bottom-right of the figure. The temporally aligned trajectories provide a set of point correspondences. We use this correspondence set to compute a 3D transformation that optimally aligns the trajectories with respect to squared Euclidean distance between each member of each correspondence. The correspondences aligned in this way have a median distance of \SI{0.62}{m} along the entire \SI{1.4}{km} trajectory. This metric is similar to the Absolute Trajectory Error (ATE) \cite{sturm12_rgbd_slam}, and is designed to show global consistency between trajectories. The ATE uses the root mean square error rather than the median, which results in an ATE of \SI{0.77}{m} for this trajectory alignment, or 0.05\% of the total trajectory length. Even with the optimal alignment of the two trajectories, the two visits to the lower right of Figure~\ref{fig:newer-traj} are visible as peaks in trajectory error, Figure~\ref{fig:newer-error}.

\subsection{Computational Performance}
\begin{figure}[t]
\centering
\input{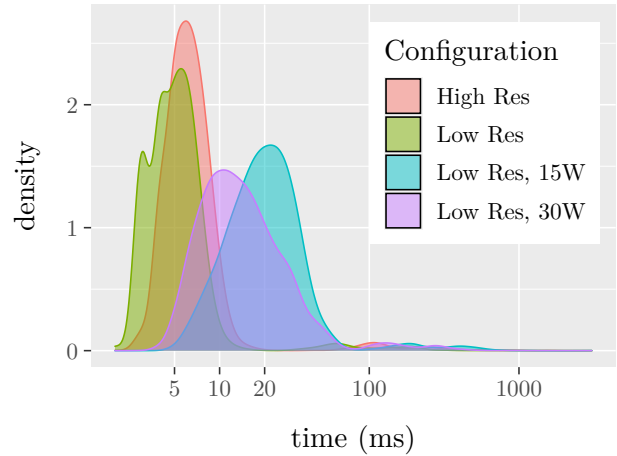}
\caption{Kernel density estimate of time to compute SLAM updates on an NVIDIA Jetson AGX Xavier at different keyframe resolutions and computer power limits.}
\label{fig:xavier-slam}
\end{figure}

The NVIDIA Jetson AGX Xavier embedded platform may be configured to run with various limits on its power consumption. These limits are observed by lowering the clock frequencies of the CPU and GPU, and disabling some of the eight CPU cores. The highest power mode runs at the maximal clock frequencies, and is limited only by the 65W power adapter. In this configuration, we consider high resolution, 2048x256, and low resolution, 1024x128, keyframes, Figure~\ref{fig:xavier-slam}. The next power limits we consider are 30W and 15W, detailed in Table~\ref{tab:xavier-power} \footnote{The 15 and 30W power levels tested are models 2 and 4 of NVIDIA's \texttt{nvpmodel} utility for the Xavier AGX.}.

\begin{table}
  \centering
  \begin{tabular}{c|c|r|r|r}
Mode & CPU & CPU & GPU & Memory \\
& Cores & MHz & MHz & MHz\\
\hline
Max & 8 & 2265.6 & 1377 & 2133\\
30W & 6 & 1450 & 900 & 1600\\
15W & 4 & 1200 & 670 & 1333\\
  \end{tabular}
  \caption{Jetson Xavier Power Modes}
  \label{tab:xavier-power}
\end{table}

The time taken to incorporate a new lidar sweep into the map is shown for each configuration in Figure~\ref{fig:xavier-slam}. The median time taken with the Xavier's maximal power budget and high resolution keyframes is \SI{6}{ms}. The figure shows a kernel density estimate of update times taken while processing the Newer College Dataset to reflect the fact that not every update takes the same amount of time: some updates trigger new keyframe creation, and some may result in relatively slow graph updates due to loop closures. This long tail of update times is more apparent when considering low resolution keyframes with maximal, 30W, and 15W power budgets, which resulted in median update times of \SI{5}{ms}, \SI{13}{ms}, and \SI{20}{ms}. Inputs are buffered when graph optimization takes longer than the interval between sweeps. The Xavier running at full power with this workload exhibits similar performance to NVIDIA's Quadro P1000 or GTX 1050 GPUs, hardware often found in laptop computers.

\section{CONCLUSIONS}

The UPSLAM mapping system has been qualitatively evaluated across diverse environments spanning indoor, outdoor, underground, aerial, and public road settings. These challenging mapping scenarios involving mud, robot collisions, dense foliage, bodies of water, and road traffic were all handled by the same mapping software adjusted with only two configuration parameters. Beyond qualitative checks for functionality, accuracy was quantified with a benchmark dataset demonstrating an absolute trajectory error of 0.05\% of the total trajectory length. Performance was measured at a median update rate of less than \SI{10}{ms} on an embedded compute platform. Workspace scaling has been demonstrated by navigating corridors and tunnels \SI{1}{m} across, as well as multi-kilometer road ways. Reconstructed geometry quality has been shown with real time visualizations supporting rich lighting effects, such as accurate shadows relating the mutual visibility of centimeter-scale geometry at ranges of hundreds of meters. We believe that this effort pushes the boundaries of robustness and performance for SLAM software, while demonstrating the effectiveness of the graph of panoramas map representation and the design choice to use all available range data during registration.

\bibliographystyle{IEEEtran}
\bibliography{../mybib/mybib.bib}

\end{document}